\documentclass[conference]{IEEEtran}

\usepackage{graphicx}

\begin{document}

\title{Visual aesthetic analysis using deep neural network: model and techniques to increase accuracy without transfer learning}

\author{\IEEEauthorblockN{Muktabh Mayank Srivastava}
\IEEEauthorblockA{ParallelDots, Inc. \\
Email: muktabh@paralleldots.com }
\and
\IEEEauthorblockN{Sonaal Kant}
\IEEEauthorblockA{ParallelDots, Inc. \\
Email: sonaal@paralleldots.com}
}


\maketitle

\begin{abstract}
We train a deep Convolutional Neural Network (CNN) from scratch for visual aesthetic analysis in images and discuss techniques we adopt to improve the accuracy. We avoid the prevalent best transfer learning approaches of using pretrained weights to perform the task and train a model from scratch to get accuracy of 78.7\% on AVA2 Dataset close to the best models available (85.6\%). We further show that accuracy increases to 81.48\% on increasing the training set by incremental 10 percentile of entire AVA dataset showing our algorithm gets better with more data.
\end{abstract}


\IEEEpeerreviewmaketitle

\section{Introduction}

Visual aesthetic analysis is the task of classifying images into being perceived as attractive or unattractive by humans. While previously handcrafted image features were a common way to solve aesthetic analysis, CNNs have recently been used to solve the problem as state of the art approaches \cite{DBLP:journals/corr/JinCPTYL16, Wang:2016:MDL:3006053.3006242, Dong2015, Mai_2016_CVPR}. They have multiple advantages such as having lesser inference time and the ability to be deployed in mobile devices after quantization. A variety of standard CNN architectures pretrained on ImageNet dataset \cite{imagenet_cvpr09} are readily available as open source for use. They are commonly utilized to achieve remarkable results on visual aesthetic datasets. However, the use of pretrained weights leaves very little scope for modifying the original architectures. On the other hand, training CNN from scratch only on a visual aesthetic dataset faces the risk of overfitting (due to smaller size of the dataset as compared to ImageNet) and limits the depth of modified architectures. The contributions of this paper are:

\begin{enumerate}
\item We propose a deep CNN architecture for visual aesthetic analysis that is specifically tailored to extract the intuitive features for underlying task. The architecture employs comparatively lesser parameters despite its depth.
\item We propose two simple tricks to train the architecture from scratch and achieve improved accuracy. First, we explore the effect of converting input images to different color spaces and find that LAB space is more sensitive towards aesthetic features of image as compared to RGB space. Second, we employ a novel training schedule, called \textbf{coherence training}, that improves accuracy further.
\end{enumerate}
With these two innovations, we are able to train our algorithm from scratch and report accuracies close to the best models even on train datasets as small as 26000 images. We also show that with increase in the amount of data, our algorithm's accuracy gets better.

 

\section{Dataset}
\label{dataset}

We used AVA2 dataset, which is a subset of original AVA dataset \cite{AVADataset}. AVA Dataset consists of 250,000 images scored for their aesthetic quality on a scale of 1 to 10. To create AVA2 dataset, the images of AVA are sorted in ascending order of their mean score and the top 10\% and bottom 10\% are taken as attractive and unattractive images respectively. Both classes of images are then split into train and test images making AVA2 a dataset of total 51,106 images, 25,553 images in both train and test. 

The model produced by training algorithm on AVA2 train set is called \textbf{model1}. We perform an additional experiment where we add to our train set more images, outside of AVA2, from 75th percentile of average score onwards and from below 25th percentile of the AVA dataset. The algorithm trained on this dataset (AVA2 train set + images from outside AVA2) is called \textbf{model2}.

\begin{figure*}[h]
  \centering
  \includegraphics[width=12cm,height=0.45\textheight]{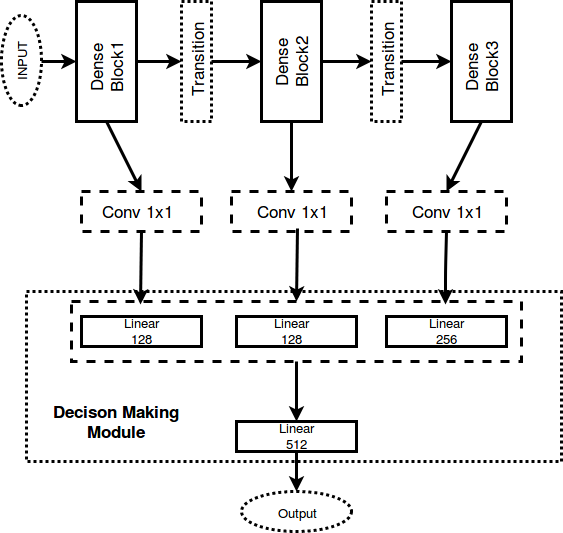}
  \caption{Our Convolutional Neural Network architecture with Dense Blocks and feature accumulation from different levels to model aesthetics}
  \label{Fig:arch}
\end{figure*}

\section{Method}
\label{method}

\subsection{Architecture}
\label{arch}
\begin{figure}[h]
  \centering
  \includegraphics[width=10cm]{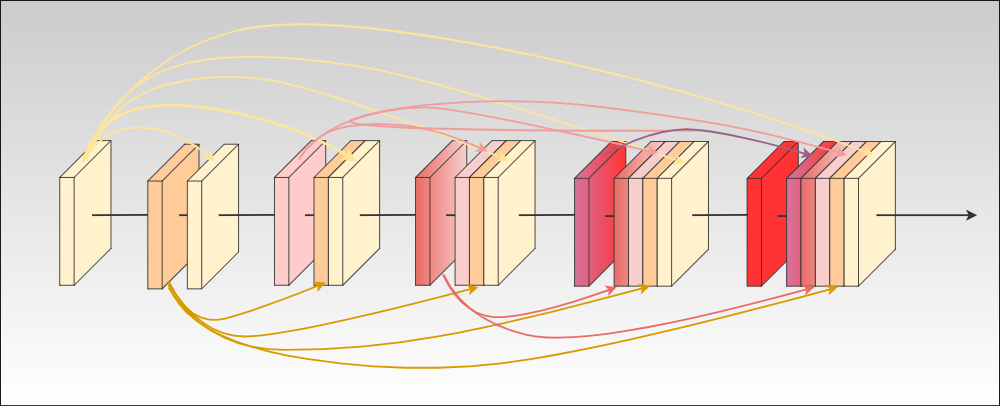}
  \caption{\textbf: DenseBlock}
  \label{Fig:densenet}
\end{figure}

\begin{figure}[h]
  \centering
  \includegraphics[width=6cm]{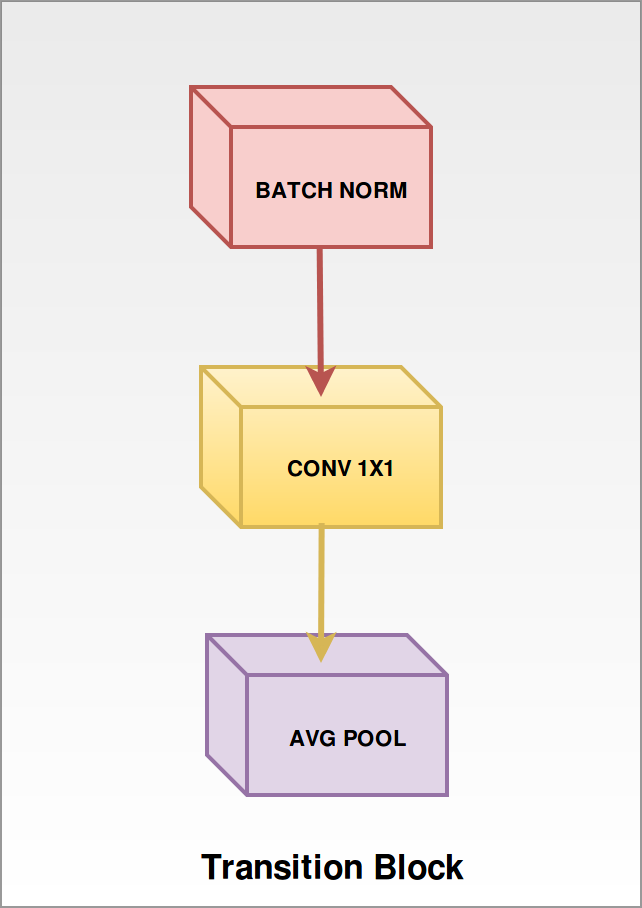}
  \caption{\textbf: Transition Block}
  \label{Fig:transitionblock}
\end{figure}

Our algorithm's architecture is inspired from ILGNet architecture \cite{DBLP:journals/corr/JinCPTYL16} in the manner that it takes both high and low level features into account to classify an image into attractive or unattractive. Figure \ref{Fig:arch} represents our proposed architecture. We use DenseNet \cite{huang2017densely} blocks in our architecture as they are known to use fewer parameters and thus avoid risk of overfitting while training from scratch. We use three Dense Blocks (A typical DenseBlock showing feature growth is represented in Figure \ref{Fig:densenet}) with growth rate of 12 in model1 and growth rate of 24 in model2. Transition Blocks (as seen in Figure \ref{Fig:transitionblock}) between Dense Blocks reduce the feature maps by half using (1x1) Convolutions. Skip connections from end of each Dense Block connect to the Decision Making module which produces final output. Each Dense Block is hypothesized to be learning to extract features at different levels. The learning at all levels is then passed as input to Decision Making Module. In the Decision Making Module, knowledge from each level is first feature-map reduced (to one-third of feature maps coming as input into the module) by using (1x1) Convolutions. Followed by (1X1) convolutions, knowledge from each level is individually transformed using Fully Connected(FC) layer. Then the output from all levels is concatenated to be passed through a FC Layer, which produces the final classification.

\subsection{Training}
\label{training}
The model was trained with SGD algorithm with momentum and decaying learning rate. We employed two simple techniques to boost the performance of our model as mentioned above. First, we converted the input images from RGB space to \textbf{LAB space}, which was found to give an accuracy boost of 1\%. Further, we used novel \textbf{coherent training} schedule for training the network from scratch. Instead of training with minibatches comprised of completely random images, we composed the minibatches with comparatively similar images related to both attractive as well as unattractive categories. The proposed technique is inspired from real-world, where learning from multiple similar examples at once leads to better understanding of the concept. In CNN framework, the approach of coherent training leads to learning of more discriminative features. We calculated the semantic representation of each image from output of pool5 layer of VGG16 network \cite{DBLP:journals/corr/SimonyanZ14a} pretrained on ImageNet dataset. We then computed nearest neighbors for each image in the aforementioned semantic space to be used as similar examples. We get an accuracy boost of 2\% by this technique.

\subsection{Testing}
\label{testing}
The model was tested on AVA2 Dataset on which the current state of the art model has an accuracy of 85.6\% \cite{DBLP:journals/corr/JinCPTYL16}. The best accuracy using hand crafted features is 68.55\% \cite{AVADataset}. In comparison, our model gets an accuracy score of 78.7\% on the AVA2 dataset (model1). When our algorithm is trained on slightly larger dataset (model2) and tested on the test set of AVA2 dataset, the accuracy reaches 81.48\%.

\subsection{Discussion}
\label{discussion_results}
The CNN architecture we propose and our training methodology are designed keeping the following points in mind :
\begin{enumerate}

\item Both low level features and high level features are important in training a CNN for visual aesthetics. This was a concept introduced by \cite{DBLP:journals/corr/JinCPTYL16}.

\item Instead of training an algorithm using a pretrained CNN on imagenet dataset as a feature extractor for the task, we train our proposed CNN architecture from scratch just on AVA2 dataset. This shows that CNNs can be trained with good accuracy on aesthetic datasets without transfer learning.

\item We introduce two training techniques, which help us get better results. First is usage of LAB space instead of RGB space as an input to our CNN.The intuition for this technique is that LAB space is designed to closely model human vision. We also use coherent learning to train CNN on minibatches that are comprised of similar images belonging to both attractive/unattractive classes. This technique is based on the intuition that when similar images from both classes are introduced in the same minibatch, the CNN is forced to learn discriminative features.

\item While we train our model on AVA2 dataset (model1) with good accuracy, we also show that the model gets better as we add more training data. We do this by adding a non-participating subset of AVA dataset into AVA2 dataset's train set and testing on AVA2 dataset's test set (model2).

\end{enumerate}

\section{Conclusion}
We present a deep CNN architecture, which can be trained from scratch on a visual aesthetic analysis dataset that gets better as we give it more data. We also propose training techniques to increase its accuracy.


\bibliography{refs}
\bibliographystyle{plain}

\end{document}